# Neural networks for dedicated neurocomputing circuits: a computational study of tolerance to noise and activation function non-uniformity

Ye min Thant[1], Methawee Nukunudompanich[1,a], Chu-Chen Chueh[2,b], Manabu Ihara[3], Sergei Manzhos[2,c*]

[1] Department of Industrial Engineering, School of Engineering, King's Mongkut Institute of Technology Ladkrabang (KMITL), 1 Chalong Krung, 1 Alley, Lat Krabang, Bangkok 10520, Thailand.

[2] Department of Chemical Engineering, National Taiwan University, Taipei 10617, Taiwan

[3] School of Materials and Chemical Technology, Institute of Science Tokyo, Ookayama 2-12-1, Meguro-ku, Tokyo 152-8552 Japan.

* Author to whom correspondence should be addressed. E-mail: manzhos.s.aa@m.titech.ac.jp , fax: +81-3-5734-3918.

## Abstract

Dedicated analog neurocomputing circuits are promising for high-throughput, low power consumption applications of machine learning (ML) and for applications where implementing a digital computer is unwieldy (remote locations; small, mobile, and autonomous devices, extreme conditions, etc.). Neural networks (NN) implemented in such circuits, however, must contend with circuit noise and the non-uniform shapes of the neuron activation function (NAF) due to the dispersion of performance characteristics of circuit elements (such as transistors or diodes implementing the neurons). We present a computational study of the impact of circuit noise and NAF inhomogeneity in function of NN architecture and training regimes. We focus on one application that requires high-throughput ML: materials informatics, using as representative problem ML of formation energies vs. lowest-energy isomer of peri-condensed hydrocarbons, formation energies and band gaps of double perovskites, and zero point vibrational energies of molecules from QM9 dataset. We show that NNs generally possess low noise tolerance with the model accuracy rapidly degrading with noise level. Single-hidden layer NNs, and NNs with larger-than-optimal sizes are somewhat more noise-tolerant. Models that show less overfitting (not necessarily the lowest test set error) are more noise-tolerant. Importantly, we demonstrate that the effect of activation function inhomogeneity can be palliated by retraining the NN using practically realized shapes of NAFs.

---

[a] E-mail: methawee.nu@kmitl.ac.th
[b] E-mail: cchueh@ntu.edu.tw
[c] E-mail: Manzhos.s.aa@m.titech.ac.jp

# 1  Introduction

Machine learning (ML) and eventually artificial intelligence (AI) can help solve problems in many areas of science and technology and beyond. ML models are typically constructed, trained, and used on digital computers. Neural networks (NN) [1] are the most widely used type of ML and are important parts of AI. When an NN is large, the CPU and RAM cost of both training the NN and its subsequent use can be high. Some of the most involved NN models, especially those trained for AI-related tasks (such as text, voice and image recognition and generation), have millions of parameters. The CPU cost of their use after training is rather high and may even have widespread societal consequences, as was recently highlighted by the recent furor surrounding the competition between DeepSeek and ChatGPT.

An NN need not be of an enormous size for the neurocomputing cost to matter. Specifically in natural sciences, NNs are widely used in ML force fields [2–4], and the cost of recall on new data, rather than that of training, can be a limiting factor [5], for example, for large length and time scale MD (molecular dynamics) [6,7]. NNs are also widely used in materials informatics for machine learning of materials properties from descriptors of chemical composition and molecular or crystal structure [8–11]. In this application, learning the descriptors-to-properties mapping is only the first step. Ultimately, one desires to discover new materials with better properties while being inexpensive, non-toxic, abundant, etc. The associated inverse problem is difficult and sometimes intractable [12–14]. Even if it is solved, one may encounter points in the space of descriptors that correspond to good properties but do not necessarily correspond to real materials. The "nearest" real material might no longer possess the desired properties.

The number of possible molecules or solid materials is astronomical and cannot be exhaustively sampled, at least at present. However, an infinitely fast ML model encoding structure-property relations would in principle obviate the need to solve the inverse problem. Instead, one could prescreen extremely large numbers of potentially synthesizable materials and select from the results. In this paradigm, a radical speedup of the recall of the NN is needed even for small NNs, and the recall cost rather than the training cost (as long as it is sufficiently small on the time scale of a project) is the critical CPU cost factor.

Such speedup can be achieved with dedicated neuromorphic circuits – hardware implementations of the NN. If it is the recall cost that is the bottleneck, the NN can potentially be trained on a digital computer and the trained model carried to a dedicated *analog* circuit. The calculation speed is then limited by the reaction time of the circuit elements, which could approach the speed of basic electronic processes (excitation, relaxation) if capacitances and inductions are minimized. To this add the advantages of lower energy consumption and compactness, that are important for applications in remote locations or in small, mobile, autonomous devices or otherwise in conditions where using a digital computer is unwieldy. The potential of neurocomputing using dedicated neuromorphic circuits is recognized and is being actively researched [15–19], and the utility of analog circuits implementing CPU-cost sensitive parts of an NN model is gaining increasing recognition [20–22].

Neurocomputing using dedicated analog circuits carries its lot of idiosyncratic issues. These include circuit noise and uncontrollable variations in parameters of circuit elements such as transistors or diodes implementing neuron activation functions (NAF). The batch-to-batch variability of semiconductor components, which can be substantial depending on the type of the semiconductor [23–26], and changes induced by the operating environment are not known at the time of NN training and can have a significant impact on the accuracy of circuit-implemented NN. Recently, some of us studied the effects of random perturbations to the shape of the NAF[27] and found that very small levels of perturbation (barely visually perceptible on the NAF plot) can severely degrade the performance of the model in the high accuracy regime (exemplified by the need to achieve correlation coefficient $R$ values between the reference and predicted data of more than 0.999), while lower-accuracy applications (with an $R$ on the order of 0.9 or less) are more robust with to such perturbations. We also studied the effects of restrictions on the type of the shape of the NAF, which may be due, e.g. to the shape of current-voltage (*I-V*) curve of used devices, but the impact of restrictions on the type of NAF shape was much less significant than that of noise. In this work, therefore, we focus on shape perturbations of the sigmoid NAF, which is often used and can be implemented in analog neurocomputing circuits [20–22].

Random noise or random perturbations of the shape of the NAF considered in Ref. [27] and elsewhere [21,22,24], however, do not reflect smooth nonuniformities among neurons due to batch-to-batch variability and environmental effects. Also, the degradation of model accuracy due to both noise and smooth non-uniformities may depend on the NN architecture, in its simplest form on the numbers of layers and neurons. In this work, we therefore study the effects of both random noise and smooth changes in the shape of the NAF what are different for different neurons, and how those effects depend on NN architecture (numbers of layers and neurons). Previous works on analog NNs considered random noise, and in the context of classification, where noise sensitivity is generally lesser [21,22,24]. We focus on regression-type problems and on one potential application of neurocomputing where both high-throughput ML and good regression accuracy are needed: materials informatics. We show that NNs generally possess low noise tolerance and that the model accuracy decreases rapidly with noise level. Single-hidden layer NNs and NNs with larger-than-optimal sizes have somewhat higher noise tolerance. Models with a lower degree of overfitting (a smaller ratio of test and training set errors, not necessarily the lowest test set error) are more robust to noise. Importantly for the perspectives of precise neurocomputing using real-life circuits, we demonstrate that the effect of activation function non-uniformity can be palliated by retraining the NN using practically realized shapes of NAFs for each neuron.

## 2 Methods

Calculations were done in Matlab R2024a and Deep Learning Toolbox™. We used custom-coded neuron activation functions to apply random or smooth noise. The *tansig* function, or hyperbolic tangent sigmoid function, is a commonly used activation function in neural networks. It is used in this work because it achieved the best performance in our tests and is realizable in analog neurocomputing circuits [20–22]. It is defined as

$$y(x) = \frac{e^x - e^{-x}}{e^x + e^{-x}}$$

(2.1)

Random perturbations simulating noise are applied as

$$y_r(x) = \frac{e^x - e^{-x}}{e^x + e^{-x}} + A(r - 0.5)$$

(2.2)

where $A$ is the amplitude and $r$ is the random number uniformly distributed on [0, 1]. $y_r(x)$ is applied during recall of the NN using a customized version of *tansig* function.

To simulate the effects of NAF shape variations due to uncontrollable variations of parameters of circuit elements such as current-voltage curves of diode or transistors implementing NAFs, smooth random functions were added to the original sigmoid NAFs,

$$y_s(x) = \frac{2}{1 + e^{-2x}} - 1 + A(Rands(x) - 0.5)$$

(2.3)

where $Rands(x)$ is a smooth random function with values distributed [0,1]. The smoothed shape is achieved with Gaussian smoothing, with a Gaussian with of 0.2 (corresponding to 1/10[th] of input range for inputs scaled on [-1, 1]) [28]. These perturbing functions are different for each neuron in each hidden layer of the NN. $y_s(x)$ is applied during recall of the NN using a customized version of *tansig* function. When retraining the NNs using realized instances of $y_s(x)$, the random seed was fixed to keep the NAFs the same during training cycles.

The calculations were performed on several representative materials informatics data sets: formation energies vs. lowest-energy isomer of peri-condensed hydrocarbons [29], formation energies and band gaps of double perovskites [30], and zero point vibrational energy (ZPVE) of molecules from QM9 dataset [31]. Relative energies of peri-condensed hydrocarbons are obtained from COMPAS-3 [29] dataset which contains about 39,000 molecules and 9,000 molecules with optimized ground state geometries and properties computed at GFN2-xTB [32] and DFT/CAM-B3LYP [33] level of theory, respectively. Among DFT-calculated properties, the formation energy vs. lowest-energy isomer is used to machine learn from the optimized structures of molecules. The descriptors were the eigenspectrum of the Coulomb's Matrix (ECM) [34] generated from the atomic coordinates of the molecules. The descriptor space was 62-dimensional since the largest hydrocarbon molecule in the datasets had 62 atoms.

Another dataset is obtained from Ref. [30] which contains band gaps and heat of formation of lead-free double perovskites materials that are promising candidates for optoelectronic devices. These properties were predicted with GBRT (gradient boosted regression tree) [35] and GPR-NN (a hybrid method of kernel regression and NN)[36] in Ref. [37] using chemical composition and structure-based features. Electronegativities, *s, p* and *d* valence orbital radii, ionization energies, distances between cations and nearest halogens and orbital energies at each site were used as descriptors, resulting in 31 features in total. The same set of features was applied in this study (we do not use the categorical variable encoding the unit cell symmetry, cubic or tetragonal, used in Ref. [30], as it was found to be unimportant in Ref. [37]).

We also machine-learn zero-point vibrational energies (ZPVE) of molecules from QM9 [31] dataset. The QM9 dataset is a subset of GDB-17 database[38] which contains 134k neutral molecules comprising with up to 9 non-hydrogen atoms (C, O, N and F). A subset of QM9 molecules with 16 atoms including hydrogen, which consists of 14,270 datapoints, is extracted to learn ZPVE from ECM features. 18 structures for which ECM could not be generated due to their inconsistent geometries are removed from the dataset, resulting in a set of 14,252 molecules. Since all molecules had 16 atoms, the dimensionality of ECM is also 16.

Single and multiple hidden layers fully connected NNs and multi-layer NN were compared. Each dataset was split into 70% for training, 10% for validation (early stopping) and 20% for the test set. For each dataset, the best NN hyperparameters including number of neurons, number of layers, epochs and learning rates were grid-searched and selected for the best test error, by averaging over 100 runs with different random training-test splits. Mean squared error (MSE) was used as performance function, and Levenberg-Marquardt (LM) algorithm [39] was used for NN parameter optimization. After identifying the best NNs with noise-free NAFs, the model was tested with various amplitudes of random noise and shape perturbation of NAFs, where a specific random train-split was fixed to enable proper comparisons. In the case of NNs using NAFs with smooth shape variations, they were also retrained using the same hyperparameters as the perturbation-free NNs.

The three dataset are used to machine-learn quantities that represent often-used targets of machine learning of materials properties, and the use of these data sets also permits comparison of the present results with those reported to date [30,37,40,41]. They also include different desired and obtainable levels of accuracy; for example, formation energies need and can often be machine learned to about 0.01 eV/atom, while band gaps need to be known to better than 0.1 eV, but that accuracy is not always obtainable [30,42]. We report most calculations for the case of formation energies vs. the lowest-energy isomer of peri-condensed hydrocarbons [29] (sections 3.1 and 3.2), and then show that similar conclusions are obtained also for the other data (section 3.3).

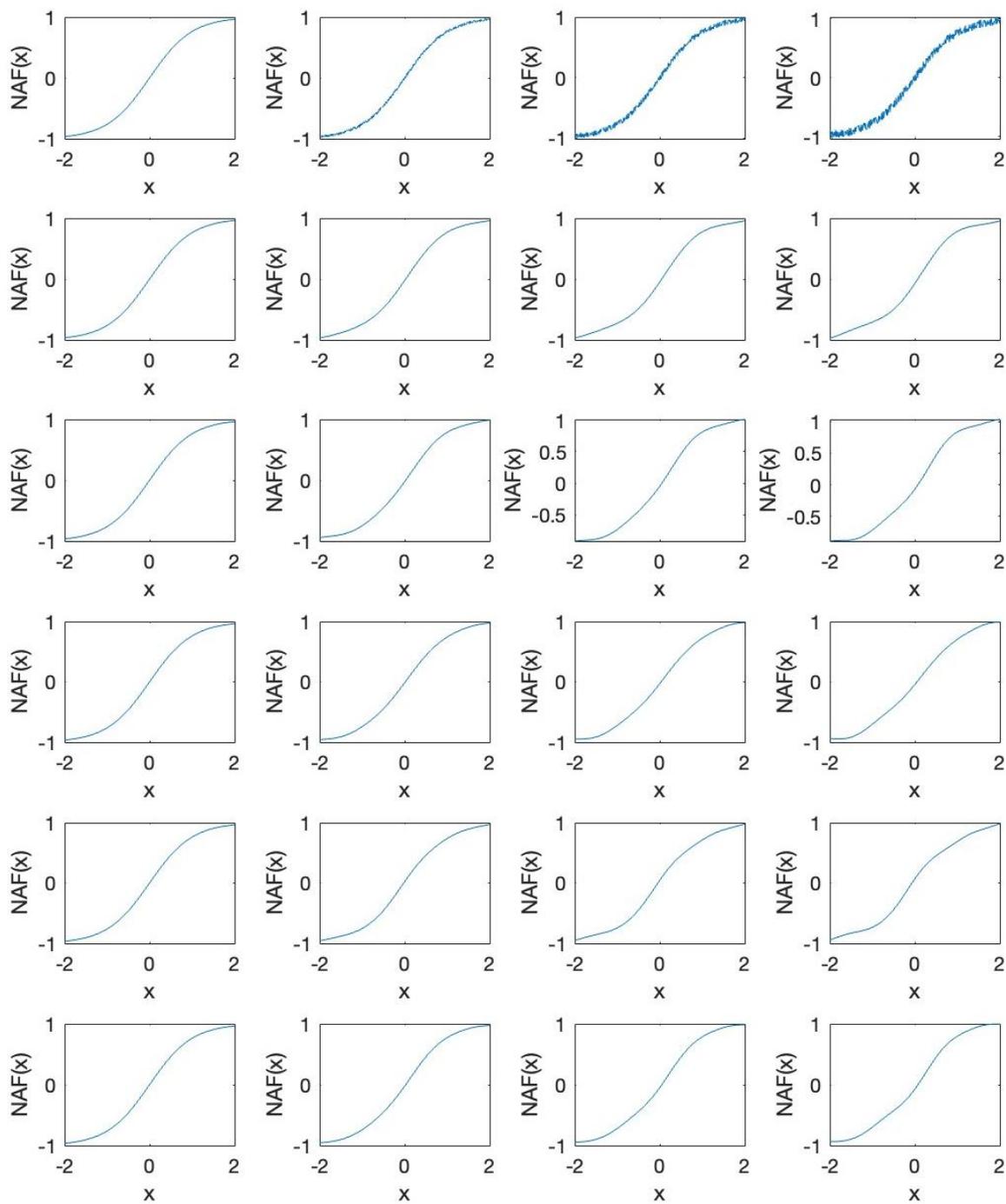

**Figure 1**. Top row: effect of noise of different amplitudes *A* (left to right: *A* = 0, 0.05, 0.10, 0.15) on the shape of a sigmoid neuron activation function. As visually the effect of random noise is similar for all neurons, one is shown. Rows 2-6: effect of NAF shape nonuniformities (smooth perturbations) with the same amplitudes of perturbation. As the effect of shape nonuniformity is different for different neurons, five NAF instances are shown.

# 3 Results

## 3.1 Effect of NAF perturbations

In **Figure 1**, we show the effect of random noise (modeling the noise in a circuit) and of smooth perturbations (modeling uncontrollable variations in the *I-V* response of semiconductor devices implementing neuron activation functions) of NAF for different magnitudes of the perturbation of the sigmoid activation function. For the case of smooth perturbation, five NAF instances are shown to demonstrate the changes of the NAF shape within a hidden layer. The magnitudes of perturbation range from minor to rather substantial, and this illustration should help to gauge the extent of the perturbation at which different levels of error are reported below. In **Figure 2** and **Figure 3**, we show the trends in training and test set errors for different magnitudes of random noise when fitting the relative energy (vs. lowest-energy isomer) of peri-condensed hydrocarbons [29] achieved with neural networks having different numbers of hidden layers (up to four) and different numbers of neurons. The numbers of neurons in different hidden layers in respective NNs are given in **Figure 2** and elsewhere in brackets, e.g. "[30]" stands for a single-hidden later NN with 30 neurons and "[15 15]" stands for a two-hidden layer NN with 15 neurons in each layer, etc. In **Figure 4** and **Figure 5**, the trends in errors are shown vs extend of smooth perturbation of the shape of the NAF. The effect of the noise and shape variations on RMSE (root mean square error) can be judged with respect to two reference error levels. One is the variance of the target which is $(0.304 \text{ eV})^2$; therefore, the fit can be said to be completely useless when the RMSE approaches this level. There is also an error level at which the error is too high for a particular application. Although this level is application-dependent, it is on the order of dozens meV in many applications for experimentally relevant predictions of stability of particular phases or isomers. 0.1 eV is chosen here as a reasonable waterline.

The best noise-free NNs approach a test set accuracy of about 0.05 eV, similar to previous reports (Ref. [40] and references therein). Without noise, with the powerful ML optimizer, there is no advantage of multilayer NN, and the best test set error is achieved with 30 neurons in a single hidden layer. With noise, we observe that larger single-hidden-layer NNs have a somewhat better noise tolerance even though they do not improve the test set RMSE in the noise-free regime. For example, a 60- or 90-neuron NN can tolerate a noise level of 0.03 before the test set error exceeds 0.1 eV. Multi-layer NNs with comparable total numbers of neurons achieve test set errors of 0.1 eV with noise levels as small as 0.015-0.025. This level of noise is barely perceptible on the NAF plot (top row of **Figure 1**). The fit becomes fully useless (test set error of 0.3 eV) when the noise level is 0.095 in a 90-neuron single hidden layer NN. Smaller single hidden layer NNs and multilayer NNs achieve this RMSE level with smaller noise magnitudes, sometimes as small as 0.045 (the [10 10 10 10] four-layer NN), i.e. they have lower noise tolerance. This superior performance of larger (than necessary in the perturbation-free regime) single-hidden-layer NNs is even more pronounced when there is a smooth NAF shape variation (exemplified in rows 2-6 of **Figure 1**): while a 60-neuron single hidden layer NN can withstand noise levels of 0.03 / 0.10 before test set RMSE reaches 0.1 / 0.3 eV, multilayer NNs produce such errors at noise levels as small as 0.01 / 0.04, respectively. **Figure 6** shows the correlation plots between the reference and model-predicted energies for the noise-free regime as well as for perturbation amplitudes of 0.05 and 0.1 for selected NN architectures, where one can appreciate the deleterious effects NAF perturbation. **Figure 6** is plotted for the case of smooth NAF shape variations; random noise perturbations result in similar plots.

In Ref. [40], we highlighted the sufficiency of a single hidden layer in many applications including the same dataset as used here. This was for the case of no perturbation and in the sense that deeper NNs do not bring a significant advantage although they may result in small improvements or at least similar performance. Here, we observe that in the presence of NAF perturbations, either random (noise) or smooth, a single layer NN is clearly advantageous in terms of tolerance to perturbation. That a larger NN is more tolerant is not surprising as it has the effect of noise averaging (as the outputs of the NAFs of the single hidden layer are simply weighted-averaged by the output weights). That a deeper NN is less tolerant must have to do with the error compounding between nonlinear layers.

*3.2  Retraining with perturbed NAF as a way to palliate the effects of NAF perturbation*

Contrary to random noise, if the perturbation of the NAF shape is smooth, one can retrain the NN with the perturbed NAFs. In real-life, this would correspond to measuring the *I-V* curves of the devices implementing the neurocomputing circuit (which would be different for different transistors or diodes and therefore for different neurons) and feeding them as NAFs for NN training to determine optimal weights and biases (corresponding to resistive elements of the circuit) under real-life NAF shapes. Note that in this scenario, each hidden layer will have different NAFs within it. In **Figure 7**, we show the results of such retraining for the cases shown in **Figure 6**. Retraining recovers most of the accuracy lost to perturbation; for example, a 30-neuron single-hidden-later NN had a test set *R* of 0.988 and test set RMSE of 0.046 eV. At a perturbation amplitude $A$ = 0.05 / 0.1, these values dropped to 0.859 / 0.697 and 0.217 / 0.427 eV, respectively, i.e. the model became practically useless. After retraining, they become 0.983 / 0.975 and 0.056 / 0.069, respectively. Even under severe NAF non-uniformities with $A$ = 0.1 (**Figure 1**), the retrained NN model still has practically usable accuracy.

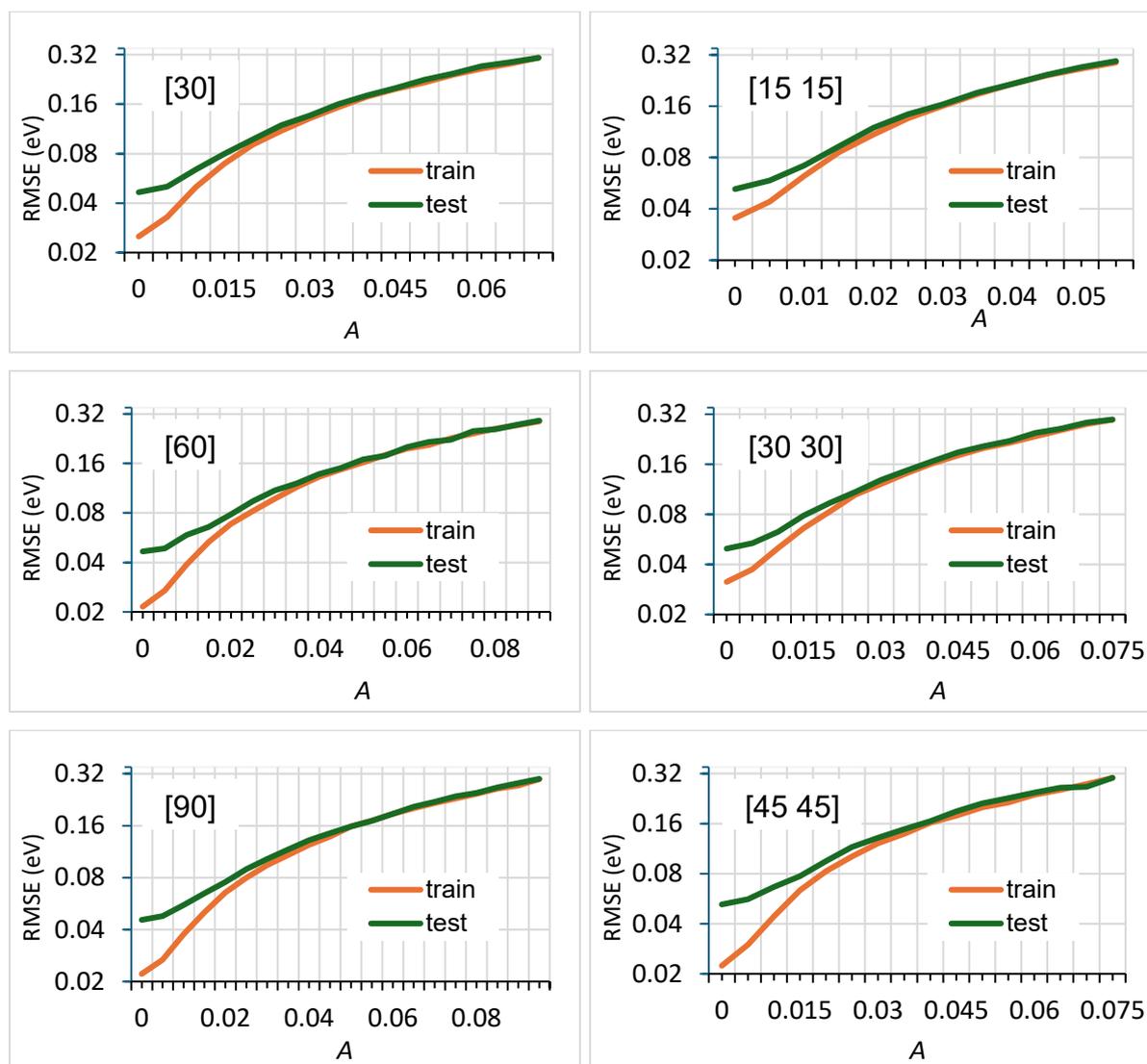

**Figure 2**. Training and test set RMSE of relative energy vs. lowest-energy isomer of peri-condensed hydrocarbons achieved with different single and two hidden layer neural networks for different magnitudes of noise. Here and in other figures, the numbers of neurons in different hidden layers are given in brackets, e.g. "[30]" stands for a single-hidden later NN with 30 neurons and "[15 15]" stands for a two-hidden layer NN with 15 neurons in each layer.

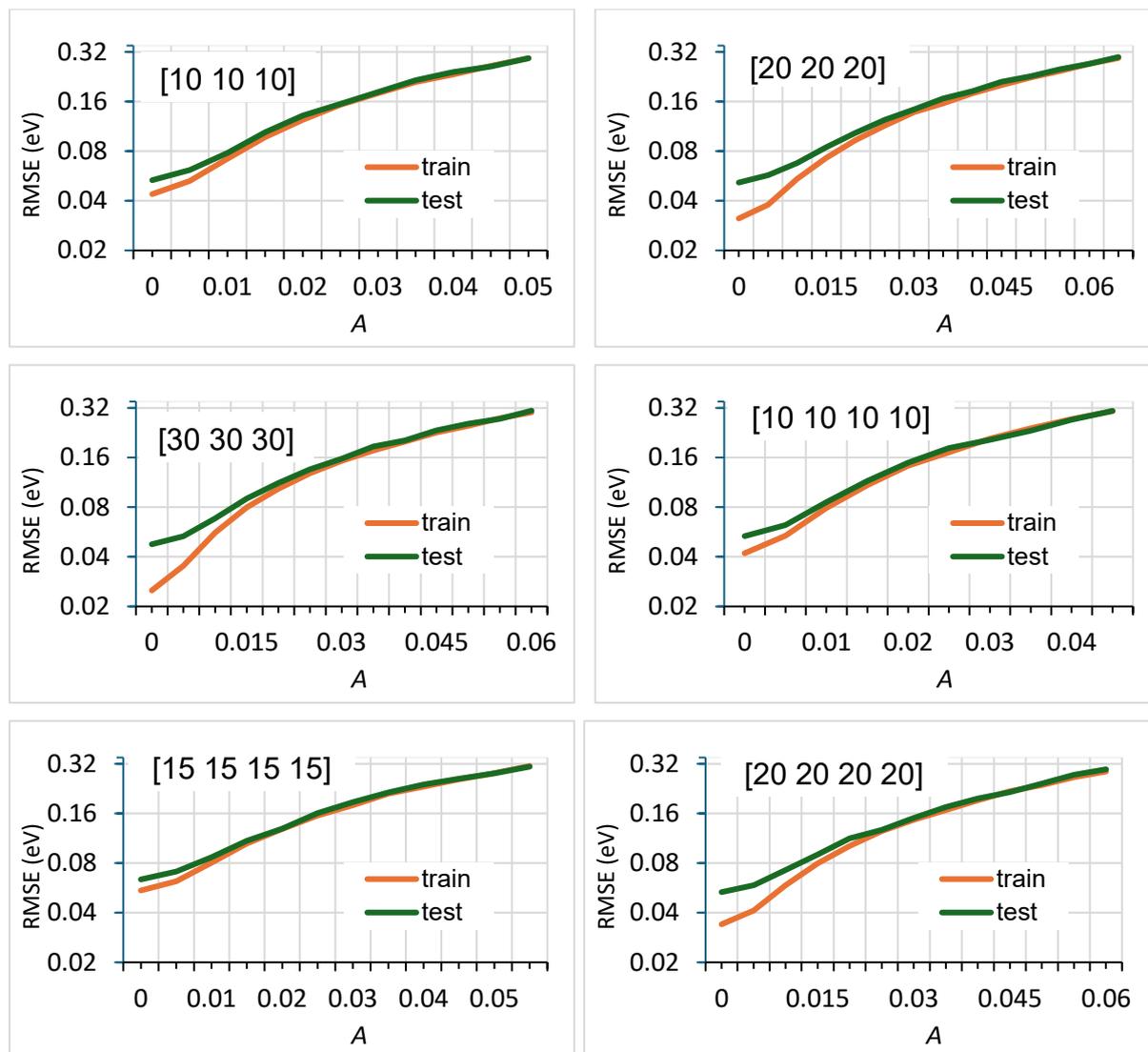

**Figure 3**. Training and test set RMSE of relative energies vs. lowest-energy isomer of peri-condensed hydrocarbons achieved with different three and four hidden layer neural networks for different magnitudes of noise. The numbers of neurons in different hidden layers are given in brackets.

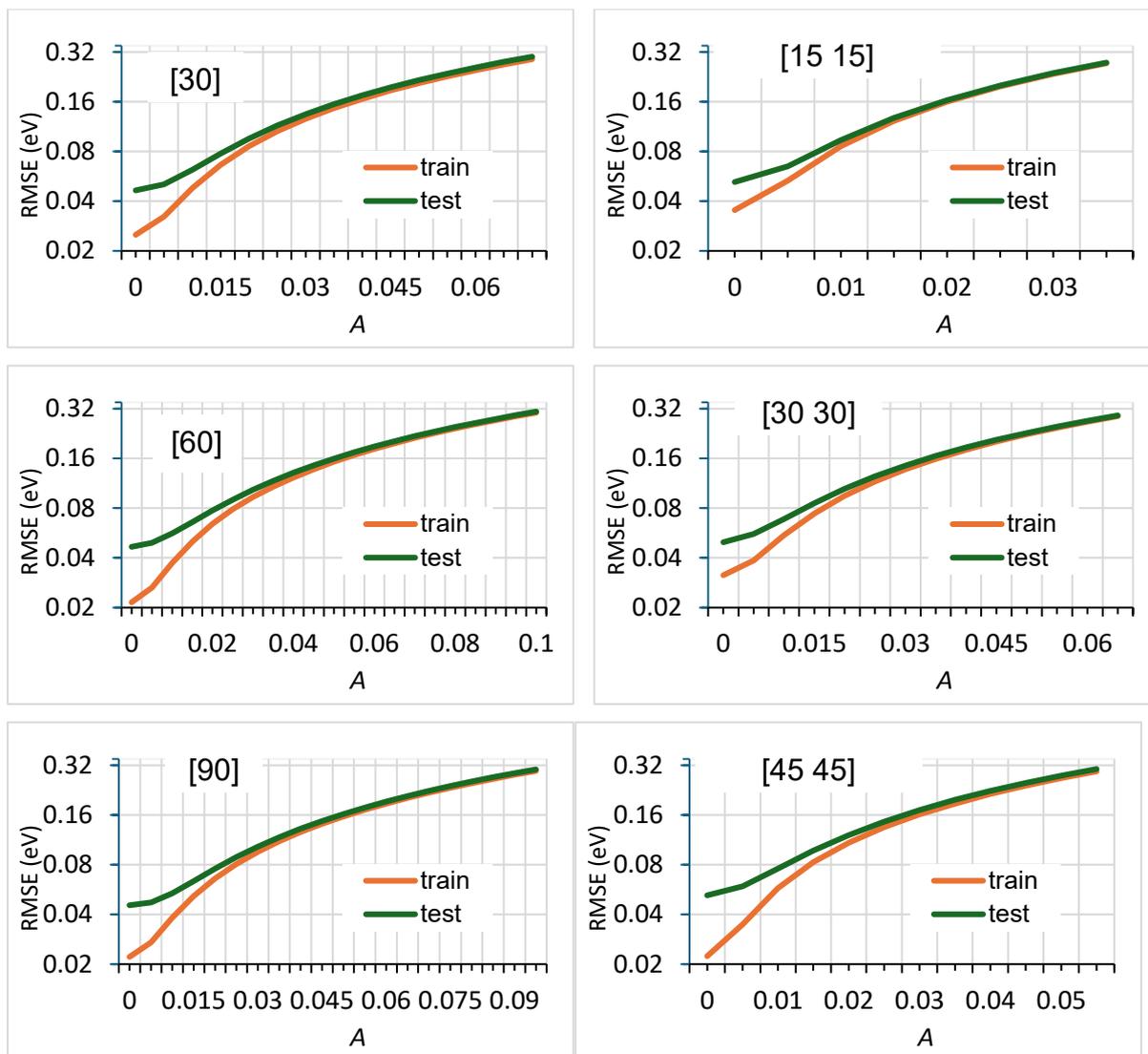

**Figure 4**. Training and test set RMSE of relative energies vs. lowest-energy isomer of peri-condensed hydrocarbons achieved with different single and two hidden layer neural networks for different magnitudes of smoothly perturbed shapes of the NAFs. The numbers of neurons in different hidden layers are given in brackets.

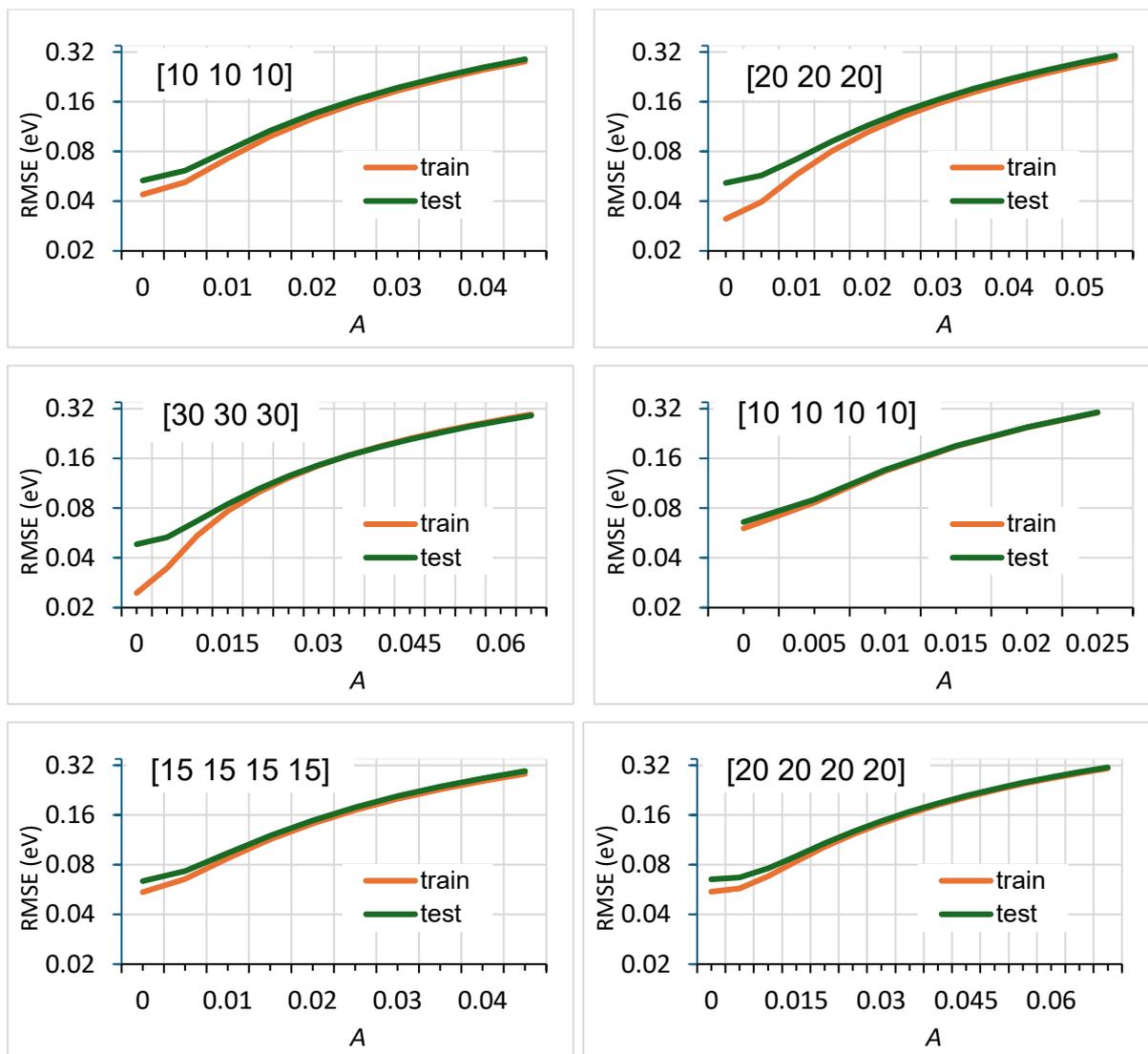

**Figure 5**. Training and test set RMSE of relative energies vs. lowest-energy isomer of peri-condensed hydrocarbons achieved with different three and four hidden layer neural networks for different magnitudes of smoothly perturbed shapes of the NAFs. The numbers of neurons in different hidden layers are given in brackets.

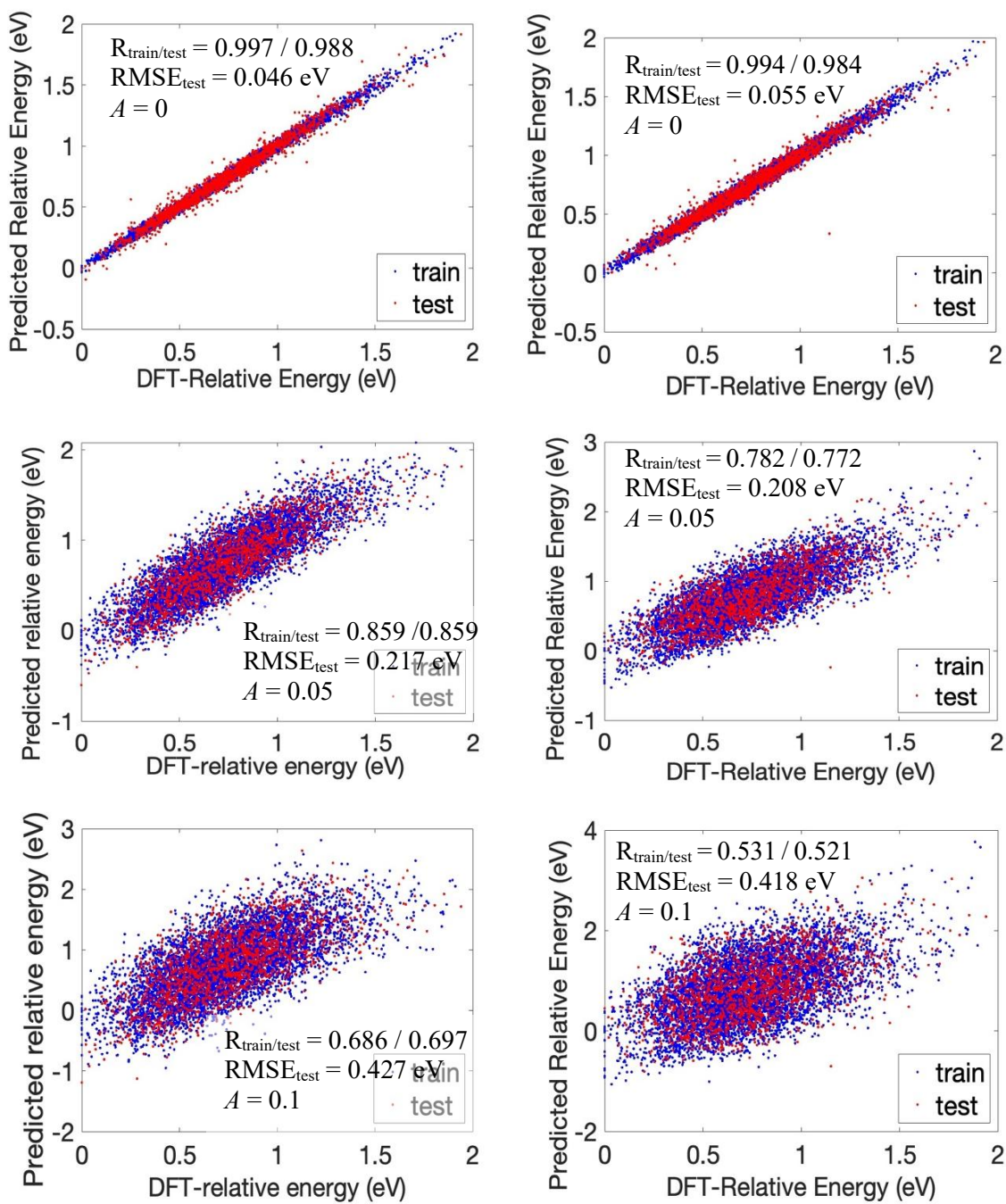

**Figure 6**. Correlation plots between the reference and NN-predicted energies for a noise-free NN (on the example of a single-hidden layer NN with 30 neurons, cf. **Figure 2**) as well as for perturbation levels $A$ of 0.05 and 0.1 with smooth NAF shape variations, for selected NN architectures, SLNN with 30 neurons (left panels) and MLNN with [20 20 20] neurons (right panels). Training set points are shown in blue and test set points in red. Training and test set correlation coefficients ($R_{train}$, $R_{test}$) and test set rmse ($RMSE_{test}$) are also shown on the plots.

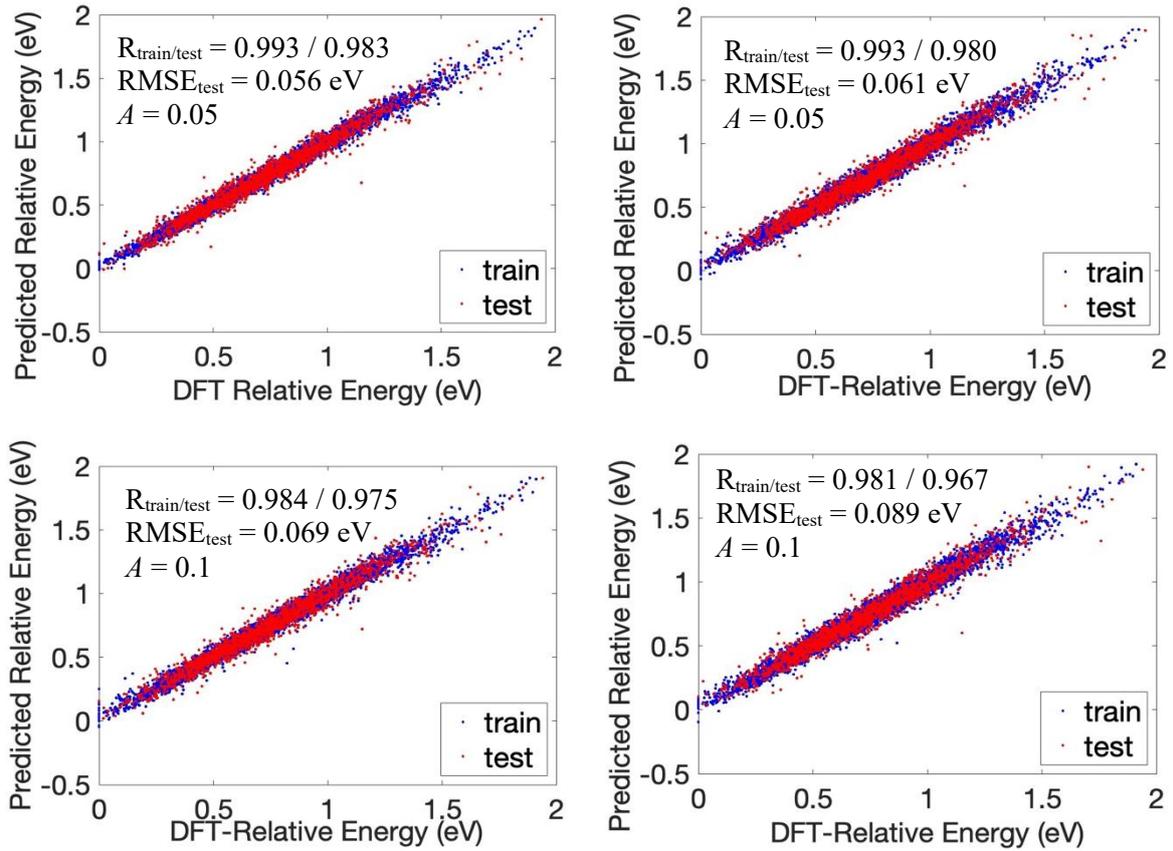

**Figure 7**. Correlation plots between the reference and NN-predicted energies for perturbation levels $A$ of 0.05 and 0.1 corresponding to smooth NAF shape variations, for selected NN architectures, SLNN with 30 neurons (left panels) and MLNN with [20 20 20] neurons (right panels) (same as shown in **Figure 6**), with NN retrained using perturbed NAF shapes. Training set points are shown in blue and test set points in red. Training and test set correlation coefficients ($R_{train}$, $R_{test}$) and test set rmse ($RMSE_{test}$) are also shown on the plots.

*3.3  Other datasets*

The above trends and conclusions are also observed with other datasets. **Figure 8** and **Figure 9** show the dependence of the training and test set errors on the amplitude of random noise or smooth shape perturbation when predicting the band gap and formation energies, respectively, of double perovskites. In the noise-free regime, the accuracy of the band gap and formation energy – with test set errors of about 0.2 eV and 0.015 eV, respectively, – is similar to the results obtained previously with various ML methods [30,37]. The figures demonstrate the deterioration of model quality and the possibility of nivellating the deleterious effect on it of random but smooth shape perturbations by retraining the NN using the realized (perturbed) NAFs. This is also demonstrated in the correlation plots showing the correlation between reference and model predicted band gap and formation energy values for a representative amplitude of the perturbation $A$.

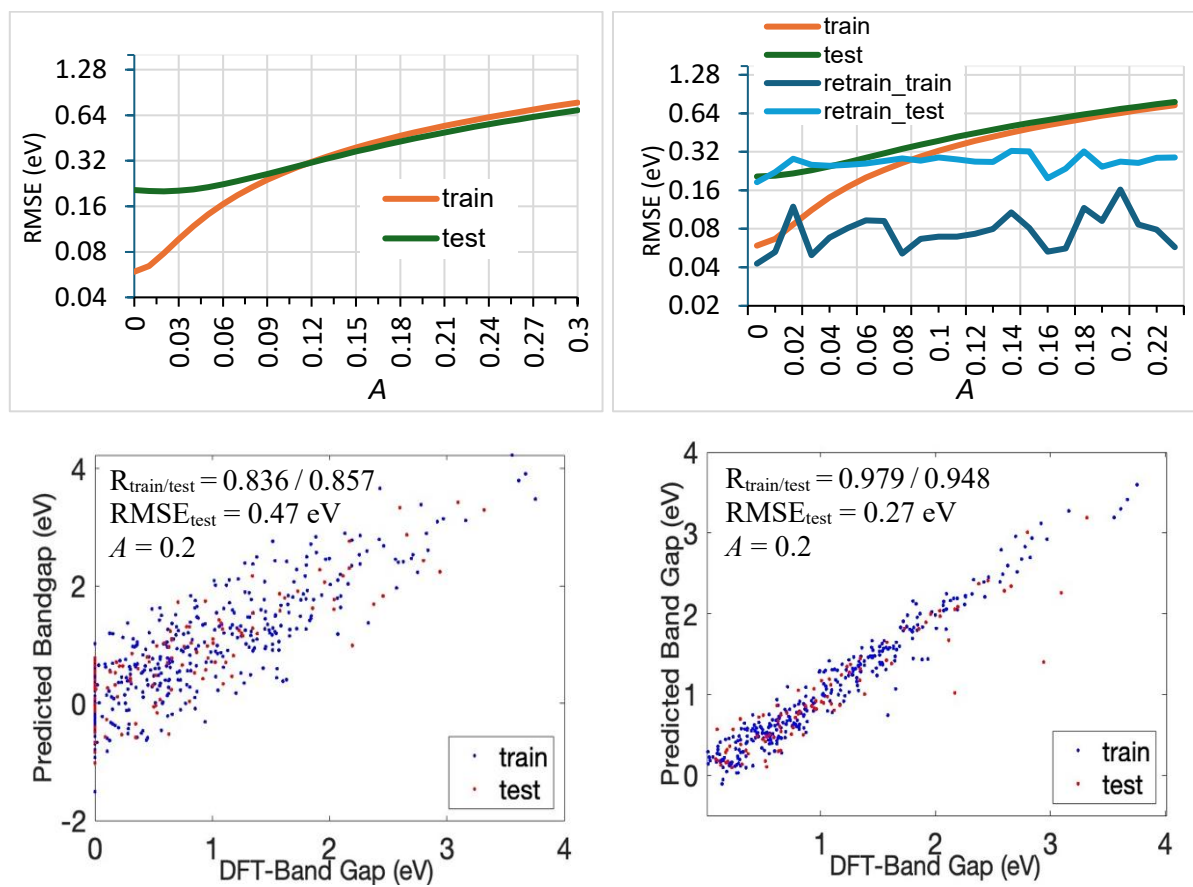

**Figure 8**. Top panels: training and test set RMSE of the band gap of lead-free double perovskites vs perturbation amplitude for the case of random (top left) and smooth (top right) NAF shape perturbation with and without retraining. Bottom panels: correlation plots for correlation between reference and model predicted band gap values under a random noise with amplitude 0.2 (bottom left) and smooth NAF shape perturbation with the same perturbation amplitude with retraining. Training set points are shown in blue and test set points in red. The optimal NN in this example had 12 neurons in the hidden layer.

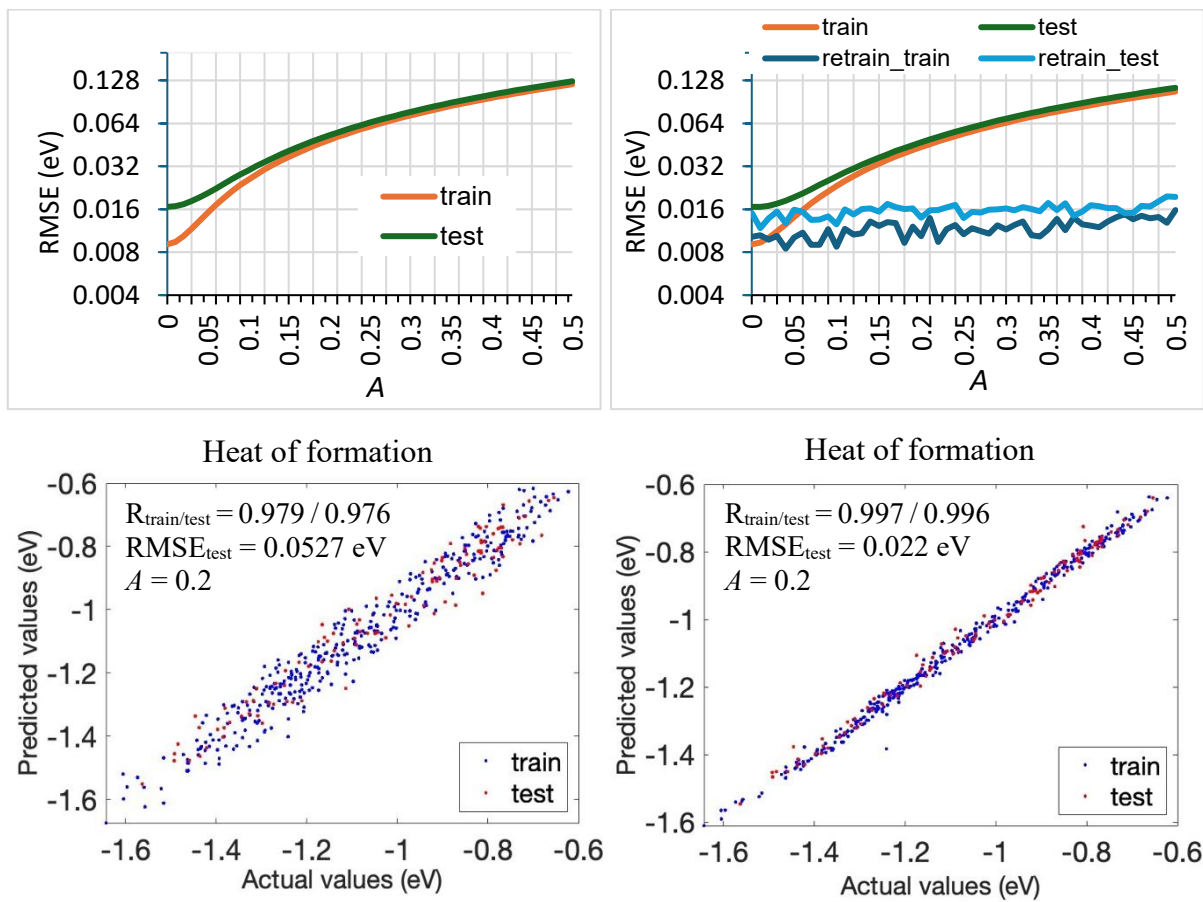

**Figure 9**. Top panels: training and test set RMSE of the formation energy of lead-free double perovskite vs perturbation amplitude for the case of random (top left) and smooth (top right) NAF shape perturbation with and without retraining. Bottom panels: correlation plots for correlation between reference and model predicted formation energies under a random noise with amplitude 0.2 (bottom left) and smooth NAF shape perturbation with the same perturbation amplitude with retraining. Training set points are shown in blue and test set points in red. The optimal NN in this example had 7 neurons in the hidden layer.

**Figure 10** shows the dependence of the training and test set errors on the amplitude of random noise or smooth shape perturbation when predicting the ZPVE for the subset of QM9 molecules containing 16 atoms. In the noise-free regime, the accuracy (test set error) of about 0.9 meV is better than that reported previously with various ML methods [41], which is expected, as we used a subset of the molecules (to have a uniform dimensionality of the ECM features of all molecules). The figure illustrates the deterioration of the quality of the ML model and the possibility of compensating for the deleterious effect of random but smooth shape perturbations by retraining the NN using the realized (perturbed) NAFs.

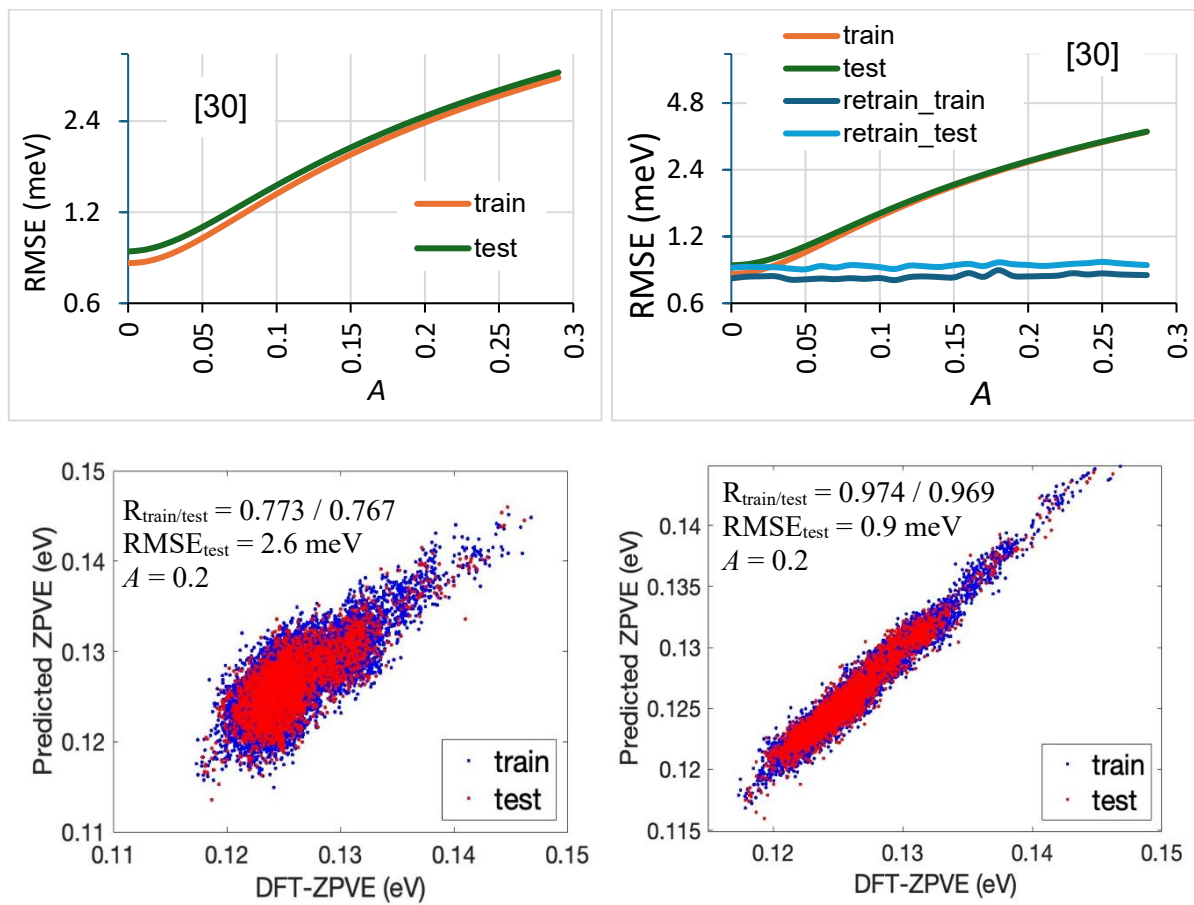

**Figure 10**. Top panels: training and test set RMSE of the zero point vibrational energy (ZPVE) of molecules from QM9 dataset vs perturbation amplitude $A$ for the case of random (top left) and smooth (top right) NAF shape perturbation with and without retraining. Bottom panels: correlation plots for correlation between reference and model predicted ZPVE under a random noise with amplitude 0.2 (bottom left) and smooth NAF shape perturbation with the same perturbation amplitude with retraining. Training set points are shown in blue and test set points in red. The optimal NN in this example had 30 neurons in the hidden layer.

## 4 Conclusions

We studied the effects of perturbations of neuron activation functions in a feed-forward artificial neural networks that are relevant for the NN implementation in analog neurocomputing circuits: random perturbations corresponding to circuit noise as well as random but smooth perturbations of the NAF shape used to model the non-uniformity of the response functions of devices implementing the NAF (such as non-uniformity of voltage-current curves corresponding to device-to-device dispersion of performance characteristics). We used materials informatics as a representative application where ultra-high throughput is desired, which is one of the prospective applications of neuromorphic circuits and considered several representative datasets that correspond to often-studied problems of machine learning of materials properties from descriptors of chemical composition and structure (machine learning of formation energies, band gaps, and ZPVE). However, these conclusions are expected to be applicable to other applications of neuromorphic computing.

We found that even minor perturbations to the NAF shape, both random noise and smooth shape variations, can substantially deteriorate the quality of the NN model. We considered how this deterioration depends on the size and architecture of the NN, and found that single-hidden-layer NNs, and NNs with larger-than-optimal sizes are somewhat more noise-tolerant. Models that show less overfitting (not necessarily the lowest test set error but a smaller ratio of test set error to training set error) are more noise tolerant.

Importantly, we showed that the realized instances of non-uniform NAFs (that could be measured once the circuit is printed) corresponding to smooth NAF perturbation can be used to retrain the NN, thereby eliminating the deleterious effect of NAF non-uniformity on the NN prediction quality. This, in our view, opens a useful approach to implementing neuromorphic circuits, especially as device-to-device NAF shape variance is substantial in some types of semiconductor devices studied for neuromorphic circuits, in particular, organic electronic devices [43–45]. The retraining with realized NAF shapes can in principle allow addressing this issue in substance.

## 5 Funding declaration

M. N. was supported by Thailand's National Science, Research and Innovation Fund (NSRF) via KMITL grant RE-KRIS/FF68/58. Y. T. was supported by the International Internship Pilot Program (IIPP) by NSTC during internship at National Taiwan University.

## 6 Authorship contribution statement

Ye min Thant: investigation, data curation, formal analysis, software, writing – review and editing.

Methawee Nukunudompanich: conceptualization, resources, supervision, project administration, writing – review and editing.

Chu-Chen Chueh: conceptualization, supervision, resources, project administration, writing – review and editing.

Manabu Ihara: supervision, writing – review and editing.

Sergei Manzhos: conceptualization, methodology, supervision, resources, project administration, writing – original draft, writing – review and editing

## 7 Declaration of Competing Interest

The author declares that they have no known competing financial interests or personal relationships that could have appeared to influence the work reported in this paper

## 8 Data availability

The data sets used in this study are available from Ref. [29] (peri-condensed hydrocarbons), Ref. [30] (double perovskites), and Ref. [31] (ZPVE). The codes are available in Supplementary Information.